\documentclass[sigconf,10pt]{acmart}

\acmYear{2025}\copyrightyear{2025}
\acmConference[ACM MOBICOM '25]{The 31st Annual International Conference on Mobile Computing and Networking}{November 4--8, 2025}{Hong Kong, China}
\acmBooktitle{The 31st Annual International Conference on Mobile Computing and Networking (ACM MOBICOM '25), November 4--8, 2025, Hong Kong, China}
\acmDOI{10.1145/3680207.3765671}
\acmISBN{979-8-4007-1129-9/25/11}

\begin{document}
\begin{sloppypar}
\title[Skyshield: Event-Driven Submillimetre Thin Obstacle Detection for Drone Flight Safety]{Poster: Skyshield: Event-Driven Submillimetre Thin Obstacle Detection for Drone Flight Safety}

\author{Zhengli Zhang$^1$, Xinyu Luo$^1$, Yucheng Sun$^2$, Wenhua Ding$^1$, \\Dongyue Huang$^{1}$, Xinlei Chen$^{1,*}$\authornote{Corresponding authors}}
\affiliation{%
  \institution{$^1$ Shenzhen International Graduate School, Tsinghua University, China}
    \country{} 
}
\affiliation{%
  \institution{$^2$ The Chinese University of Hong Kong, Shenzhen, China}
  \country{} 
}

\renewcommand{\authors}{Zhengli Zhang, Xinyu Luo, Yucheng Sun, Wenhua Ding, Dongyue Huang, Xinlei Chen}

\renewcommand{\shortauthors}{Zhengli Zhang, et al.}

\begin{abstract}

Drones operating in complex environments face a significant threat from  thin obstacles, such as steel wires and kite strings at the submillimeter level, which are notoriously difficult for conventional sensors like RGB cameras, LiDAR, and depth cameras to detect. This paper introduces SkyShield, an event-driven, end-to-end framework designed for the perception of submillimeter scale obstacles. Drawing upon the unique features that thin obstacles present in the event stream, our method employs a lightweight U-Net architecture and an innovative Dice-Contour Regularization Loss to ensure precise detection. Experimental results demonstrate that our event-based approach achieves mean F1 Score of 0.7088 with a low latency of 21.2 ms, making it ideal for deployment on edge and mobile platforms.

\end{abstract}


\maketitle
\section{Introduction}
Drones are being widely adopted for various applications, but their deployment in complex environments is challenged by the critical hazard of submillimetre-thin obstacles. In 2025 alone, over 10 small drone crashes were caused by kite strings, highlighting the urgency. Traditional methods, like motion estimation techniques \cite{Wang2020UAVEP}, struggle with these obstacles due to poor visual saliency and motion artifacts.

\begin{figure}[t!]
\centering
\setlength{\abovecaptionskip}{0.3em}
\includegraphics[width=0.9\linewidth]{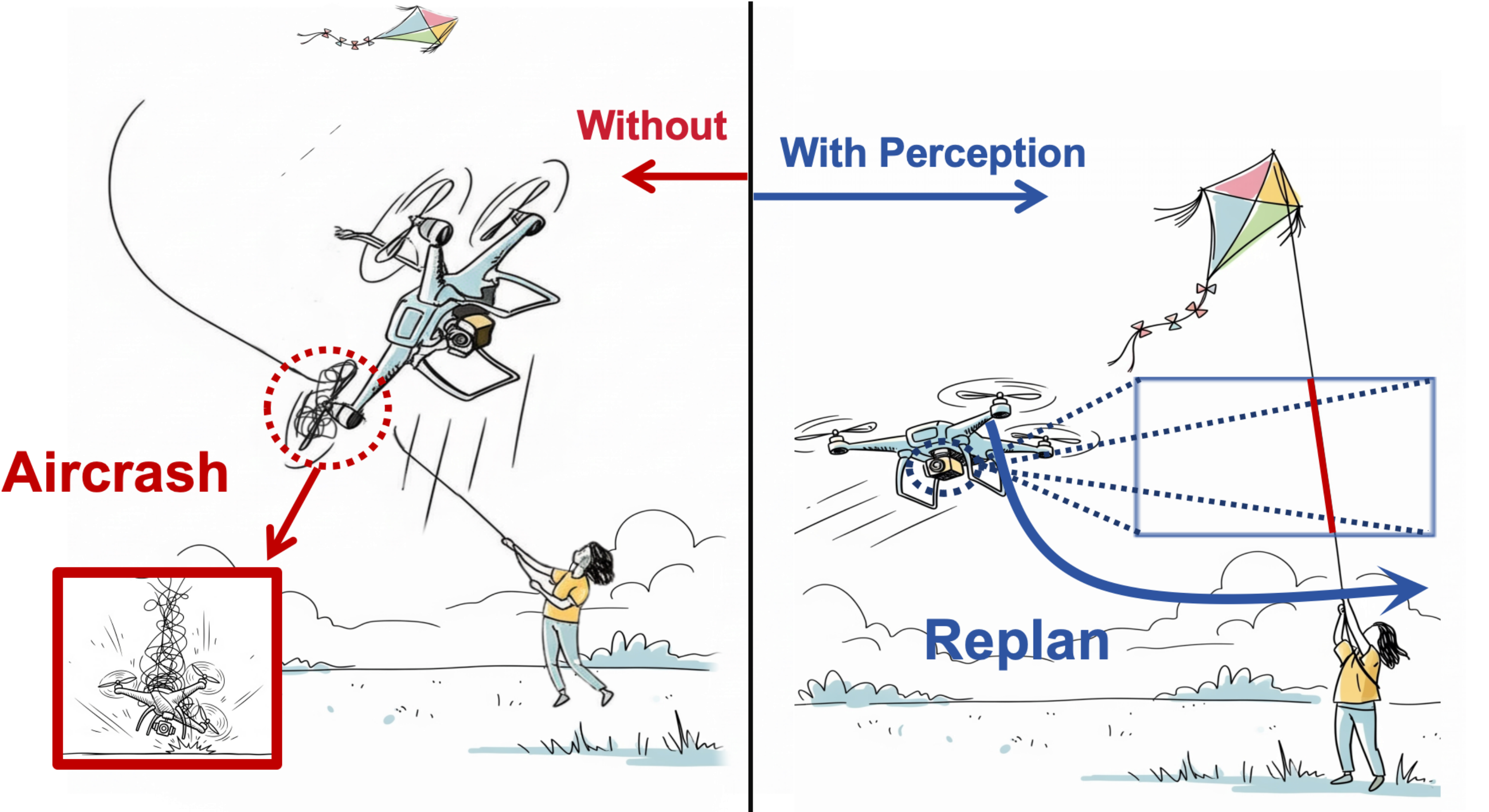}
\caption{The importance of Submillimetre thin obstacle detection for drones.}
\label{fig:introfigure}
\vspace{-0.3cm}
\end{figure}

\begin{figure*}[t!] 
\centering
\setlength{\abovecaptionskip}{0.3em}
\includegraphics[width=0.95\textwidth]{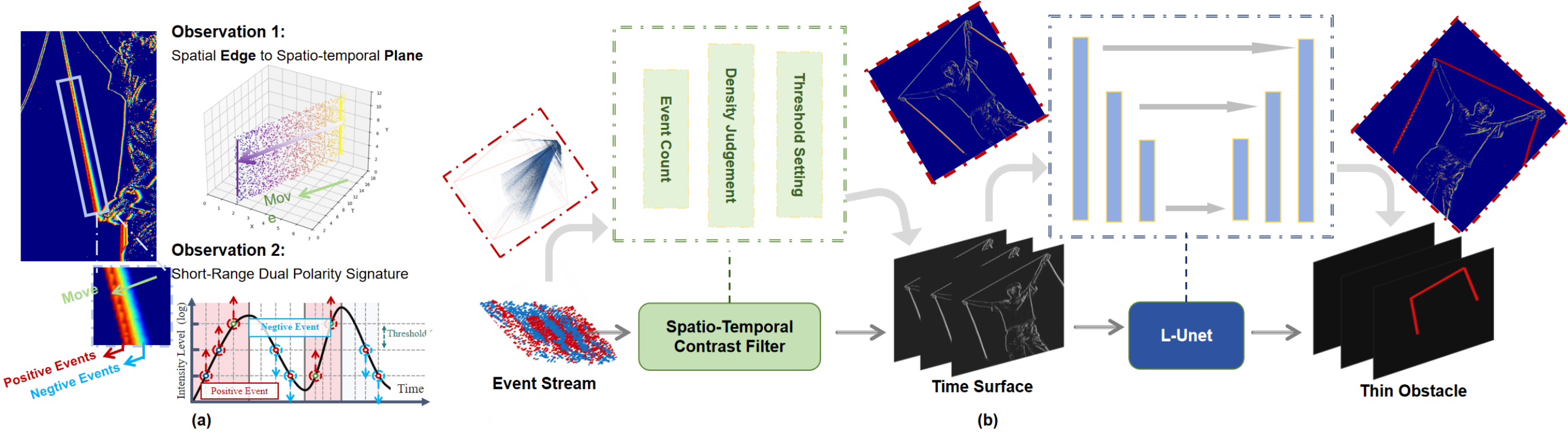} 
\vspace{-0.2cm}
\caption{Observations of Submillimetre thin obstacles in event and the skyshield workflow}
\label{fig:system_network}
\vspace{-0.1cm}
\end{figure*}

\textbf{Submillimetre Thin Obstacle Detection with event camera.}
To address the limitations of conventional sensors, we employ an event camera, which asynchronously records pixel-level brightness changes with exceptional temporal resolution\cite{wang2025towards}. Indeed, the integration of event cameras onto drones has recently been explored for applications such as obstacle avoidance \cite{xu2023biodrone} , aerial object localization \cite{wang2025mme, 10.1145/3636534.3694721}, and denoise\cite{ruan2025premamba4dstatespace}. 
Building on this promising approach, we leverage the unique capability of event cameras to transform linear structures into detectable time surfaces, offering a highly promising solution for perceiving extremely thin obstacles.


Despite its potential, translating this intuition into a practical detection system presents two key technical challenges:

\noindent $\bullet$ \textbf{Disambiguating from Background Edges.} Event cameras indiscriminately detect all edges, making it difficult to distinguish a line from overwhelming background clutter.

\noindent $\bullet$ \textbf{Managing the Data Deluge.} The immense data rate (up to $10^6$ Hz) creates a significant computational bottleneck for real-time signal extraction.

\textbf{Our work.}
We introduce Skyshield, a lightweight, event-based framework for real-time submillimeter thin obstacle detection. To our knowledge, it's the first end-to-end method to achieve this challenging task with a low latency of 21ms. 



\section{System Design}

Skyshield is a two-module system for drone safety using event cameras (Fig. \ref{fig:system_network}(b)). First, an Event Data Preprocessing module filters the event stream into a dense Time Surface. Second, a lightweight Line U-shaped CNN called LUnet processes this Time Surface to output a heatmap of thin line features. To ensure accurate and precise predictions, LUnet uses a custom Dice-Contour Regularization Loss.

\vspace{-0.4cm}
\subsection{Observation}
Our approach is driven by two fundamental insights into why the event domain is exceptionally suited for this detection task, as illustrated in Fig. \ref{fig:system_network}(a).

\textit{\textbf{Observation 1: Dimensionality Expansion}}
A thin obstacle, which appears as a sparse line in spatial domain, is transformed by an event camera's motion capture into a dense 2D surface within the spatio-temporal (x, y, t) domain. This expansion makes the feature more voluminous.

\textit{\textbf{Observation 2: Short-Range Dual Polarity}}
Due to an event camera's polarized nature, a moving thin obstacle generates a distinctive dual polarity signature. Its leading edge creates a band of positive events, while its trailing edge simultaneously creates an adjacent band of negative events. Because the obstacle is so thin, these two bands are generated in close spatial and temporal proximity.

\vspace{-0.3cm}

\subsection{Event Data Preprocessing}
The Event Data Preprocessing module performs two key transformations. First, it uses a spatio-temporal contrast (STC) filter to remove noise from the raw event stream, reducing the event rate by over 80\%. It then converts these filtered, asynchronous events into a dense, continuous-valued Time Surface. This is achieved by mapping events to a per-pixel timestamp buffer $T(x,y)$ and normalizing them with an exponential decay, as defined in the equation below, where $t_{ref}$ is the current time and $\tau$ is a decay constant. The resulting floating-point image aggregates events into a compact feature map that preserves fine temporal gradients, which serves as the input for our analysis.
\begin{equation}
    S(x,y) = exp(-\frac{t_{\rm{ref}}-T(x,y)}{\tau } )
\end{equation}

\subsection{Line Feature Extraction with LUnet}

The lightweight Line U-shaped CNN(LUnet) for efficient line feature extraction, combining encoder-decoder architecture with skip connections to capture contextual and spatial details from Time Surface data.

To leverage unique bilateral event pattern of submillimeter thin obstacles, we propose a new loss function: the Dice-Contour Regularization Loss. 
This loss consists of two components: a Dice Loss to ensure accurate location of lines, and a Contour Regularization Term to encourage network to output precise, thin centerlines by penalizing thick predictions.

The complete loss function is defined as:
\begin{align}
    L_{\text{total}} &= L_{\text{Dice}} + \lambda L_{\text{reg}} \label{eq:total_loss} \\
    L_{\text{Dice}} &= 1 - \frac{2 \sum_{i} p_i g_i + \epsilon}{\sum_{i} p_i + \sum_{i} g_i + \epsilon} \label{eq:dice_loss} \\
    L_{\text{reg}} &= \left( 1 - \frac{P(p) + \delta}{2 \cdot A(p) + \delta} \right)^2 \label{eq:reg_loss}
\end{align}

$L_{\text{Dice}}$ is the standard Dice Loss, used to measure the overlap between the predicted heatmap and the ground truth. $L_{\text{reg}}$ is the contour regularization term, which is designed to encourage a high perimeter-to-area ratio, a key geometric characteristic of thin lines. 
$p_i$ and $g_i$ are the predicted probability and ground truth label for pixel $i$, respectively. $\lambda$ is a hyperparameter used to balance the weights of the two loss terms. $A(p)$ and $P(p)$ represent the area and perimeter of the predicted line, respectively. $\epsilon$ and $\delta$ are small smoothing terms to prevent division by zero.
\begin{figure}[t!]
    \setlength{\abovecaptionskip}{0.3cm} 
    \centering
    \includegraphics[width=0.8\linewidth]{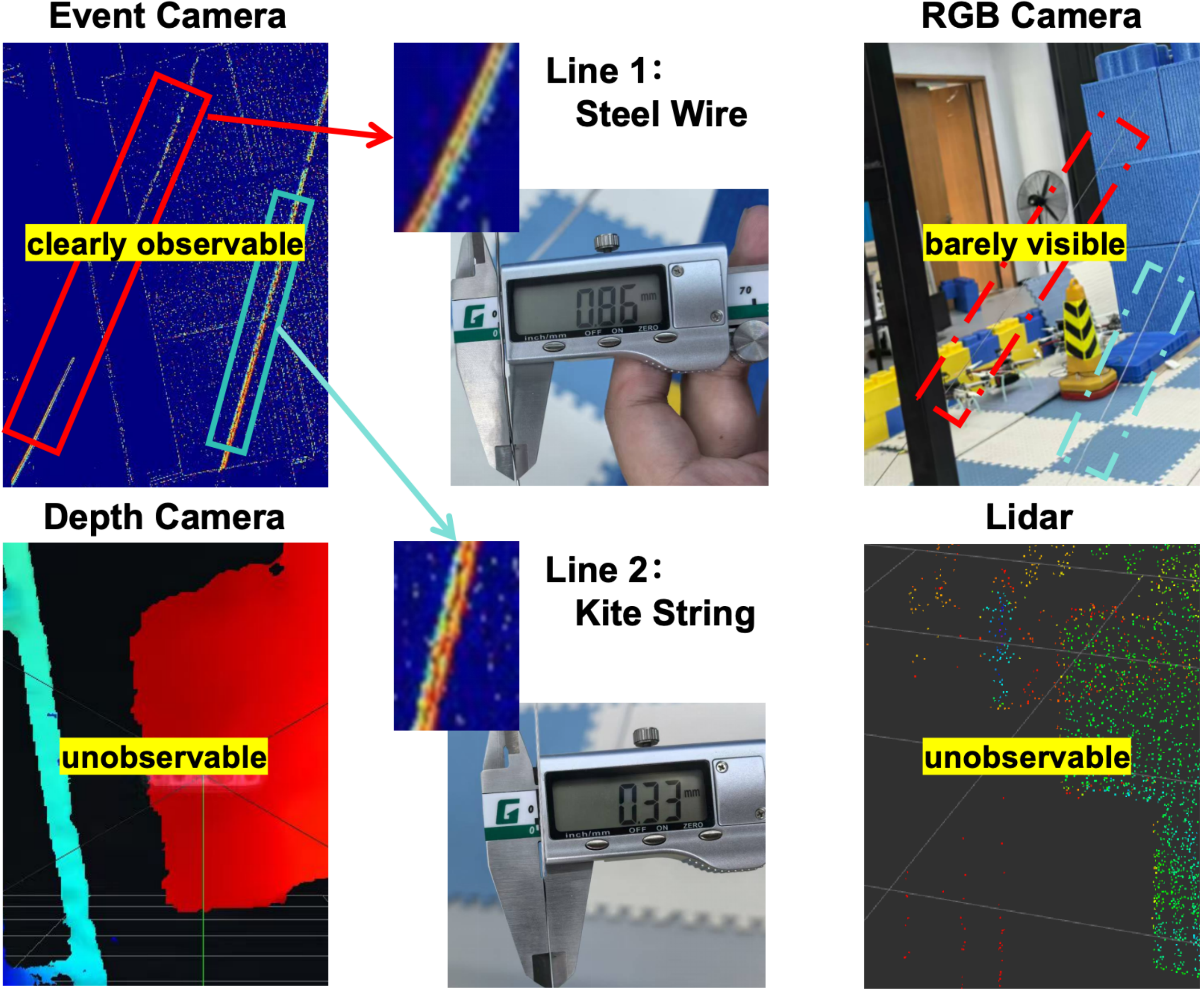}
    \caption{Comparison of perception capabilities of different sensors for submillimeter thin obstacles}
    \label{fig:diffsenser}
    \vspace{-0.3cm}
\end{figure}

\section{Evaluation}
\label{sec:evaluation}

\subsection{Experimental Setup}

\textbf{Setup.}
Tested our lightweight, real-time deployment design on NVIDIA Jetson Orin NX, using Prophesee EVK4 HD event camera to build a submillimeter-thin obstacle detection dataset with 0.86mm steel wire and 0.33mm kite string under various indoor and outdoor conditions.

\textbf{Baseline.}
To evaluate the performance of our proposed Skyshield, we compared it against two traditional line detection baselines: Hough Transform, which detects lines using a voting procedure, and the efficient Line Segment Detector (LSD) \cite{4731268}, which operates without the need for parameter tuning.

\textbf{Metrics.}
We evaluated performance using three metrics:
$(i)$  Intersection over Union (IoU) for overlap accuracy,
$(ii)$ Dice coefficient (equivalent to F1-score) for small-object sensitivity,
$(iii)$ Inference Time for computational efficiency.
As shown in Table \ref{tab:evaluation}, our method outperforms traditional approaches in all metrics.



\subsection{Multi-sensor Comparison}
To evaluate sensor performance on fine threads, we conducted a multi-sensor comparison using a 0.86mm steel wire and a 0.33mm kite string. As shown in Fig. \ref{fig:diffsenser}, depth cameras (D435i) and LiDAR (Mid360) completely failed to detect the threads. The high-resolution RGB camera (3840x2160) could only capture faint features that were easily obscured by background clutter.

In contrast,the event camera(Prophesee EVK4 1280*720) clearly detected the threads, even after filtering. This observation highlights its superior sensitivity for fine-thread detection, demonstrating that event-based vision is a more viable solution compared to conventional sensing modalities.

\subsection{Overall Performance}

\begin{table}[h]
\setlength{\abovecaptionskip}{0.1cm} 
\setlength{\belowcaptionskip}{0.1cm}
    \centering
    \small
    \begin{tabular}{lcccc}
        \toprule
        \textbf{Method} & \textbf{Mean IoU} & \textbf{Mean Dice} &  \textbf{Mean Inference Time} \\
        \midrule
        \textbf{LUnet} & \textbf{0.5704} & \textbf{0.7088} & \textbf{21.2 ms} \\
        HOUGH & 0.0694 & 0.1167 & 26.6 ms \\
        LSD & 0.0768 & 0.1344  & 33.0 ms \\
        \bottomrule
    \end{tabular}
    \caption{Performance comparison of LUnet with traditional baseline methods.}
    \label{tab:evaluation}
    \vspace{-0.5cm}
\end{table}

\label{sec:analysis}

Skyshield demonstrates superior performance across all key metrics. With a mean Dice Coefficient of \textbf{0.7088}, it significantly outperforms baselines and accurately thin lines. Its average inference time of just \textbf{21.2ms} also makes it the fastest method, providing a combination of high accuracy and speed ideal for real-time drone safety applications.

\section{Conclusion}
Skyshield is a novel event-driven system that uses a lightweight Line U-Net (LUnet) and a novel Dice-Contour Regularization Loss to detect submillimeter thin obstacles in real-time. Evaluation demonstrates superior accuracy and speed compared to traditional methods, making it an efficient solution for enhancing drone safety on edge platforms.

\section*{Acknowledgement} 
This paper was supported by the Natural Science Foundation of China under Grant 62371269 and Meituan Academy of Robotics Shenzhen.

\bibliographystyle{ACM-Reference-Format}
\bibliography{reference}


\begin{thebibliography}{7}


\ifx \showCODEN    \undefined \def \showCODEN     #1{\unskip}     \fi
\ifx \showISBNx    \undefined \def \showISBNx     #1{\unskip}     \fi
\ifx \showISBNxiii \undefined \def \showISBNxiii  #1{\unskip}     \fi
\ifx \showISSN     \undefined \def \showISSN      #1{\unskip}     \fi
\ifx \showLCCN     \undefined \def \showLCCN      #1{\unskip}     \fi
\ifx \shownote     \undefined \def \shownote      #1{#1}          \fi
\ifx \showarticletitle \undefined \def \showarticletitle #1{#1}   \fi
\ifx \showURL      \undefined \def \showURL       {\relax}        \fi
\providecommand\bibfield[2]{#2}
\providecommand\bibinfo[2]{#2}
\providecommand\natexlab[1]{#1}
\providecommand\showeprint[2][]{arXiv:#2}

\bibitem[Wang et~al\mbox{.}(2020)]%
        {Wang2020UAVEP}
\bibfield{author}{\bibinfo{person}{Dashuai Wang}, \bibinfo{person}{Wei Li}, \bibinfo{person}{X. Liu}, \bibinfo{person}{Nan Li}, {and} \bibinfo{person}{Chunlong Zhang}.} \bibinfo{year}{2020}\natexlab{}.
\newblock \showarticletitle{UAV environmental perception and autonomous obstacle avoidance: A deep learning and depth camera combined solution}.
\newblock \bibinfo{journal}{\emph{Comput. Electron. Agric.}}  \bibinfo{volume}{175} (\bibinfo{year}{2020}), \bibinfo{pages}{105523}.
\newblock
\urldef\tempurl%
\url{https://api.semanticscholar.org/CorpusID:225445299}
\showURL{%
\tempurl}


\bibitem[Wang et~al\mbox{.}(2025)]%
        {wang2025towards}
\bibfield{author}{\bibinfo{person}{Haoyang Wang}, \bibinfo{person}{Ruishan Guo}, \bibinfo{person}{Pengtao Ma}, \bibinfo{person}{Ciyu Ruan}, \bibinfo{person}{Xinyu Luo}, \bibinfo{person}{Wenhua Ding}, \bibinfo{person}{Tianyang Zhong}, \bibinfo{person}{Jingao Xu}, \bibinfo{person}{Yunhao Liu}, {and} \bibinfo{person}{Xinlei Chen}.} \bibinfo{year}{2025}\natexlab{}.
\newblock \showarticletitle{Towards Mobile Sensing with Event Cameras on High-agility Resource-constrained Devices: A Survey}.
\newblock \bibinfo{journal}{\emph{arXiv preprint arXiv:2503.22943}} (\bibinfo{year}{2025}).
\newblock


\bibitem[Xu et~al\mbox{.}(2023)]%
        {xu2023biodrone}
\bibfield{author}{\bibinfo{person}{Jingao Xu}, \bibinfo{person}{Danyang Li}, \bibinfo{person}{Zheng Yang}, \bibinfo{person}{Yishujie Zhao}, \bibinfo{person}{Hao Cao}, \bibinfo{person}{Yunhao Liu}, {and} \bibinfo{person}{Longfei Shangguan}.} \bibinfo{year}{2023}\natexlab{}.
\newblock \showarticletitle{Taming Event Cameras with Bio-Inspired Architecture and Algorithm: A Case for Drone Obstacle Avoidance}. In \bibinfo{booktitle}{\emph{Processings of the 29th ACM MobiCom}}.
\newblock


\bibitem[Wang et~al\mbox{.}({[n.\,d.]})]%
        {wang2025mme}
\bibfield{author}{\bibinfo{person}{Haoyang Wang}, \bibinfo{person}{Jingao Xu}, \bibinfo{person}{Xinyu Luo}, \bibinfo{person}{Ting Zhang}, \bibinfo{person}{Xuecheng Chen}, \bibinfo{person}{Ruiyang Duan}, \bibinfo{person}{Jialong Chen}, \bibinfo{person}{Yunhao Liu}, \bibinfo{person}{Jianfeng Zheng}, \bibinfo{person}{Weijie Hong}, {et~al\mbox{.}}} \bibinfo{year}{[n.\,d.]}\natexlab{}.
\newblock \showarticletitle{mmE-Loc: Facilitating Accurate Drone Landing with Ultra-High-Frequency Localization}.
\newblock \bibinfo{journal}{\emph{arXiv preprint arXiv:2507.09469}} (\bibinfo{year}{[n.\,d.]}).
\newblock


\bibitem[Luo et~al\mbox{.}(2024)]%
        {10.1145/3636534.3694721}
\bibfield{author}{\bibinfo{person}{Xinyu Luo}, \bibinfo{person}{Haoyang Wang}, \bibinfo{person}{Ciyu Ruan}, \bibinfo{person}{Chenxin Liang}, \bibinfo{person}{Jingao Xu}, {and} \bibinfo{person}{Xinlei Chen}.} \bibinfo{year}{2024}\natexlab{}.
\newblock \showarticletitle{EventTracker: 3D Localization and Tracking of High-Speed Object with Event and Depth Fusion}. In \bibinfo{booktitle}{\emph{Proceedings of the 30th Annual International Conference on Mobile Computing and Networking}} (Washington D.C., DC, USA) \emph{(\bibinfo{series}{ACM MobiCom '24})}. \bibinfo{publisher}{Association for Computing Machinery}, \bibinfo{address}{New York, NY, USA}, \bibinfo{pages}{1974–1979}.
\newblock
\showISBNx{9798400704895}
\href{https://doi.org/10.1145/3636534.3694721}{doi:\nolinkurl{10.1145/3636534.3694721}}


\bibitem[Ruan et~al\mbox{.}(2025)]%
        {ruan2025premamba4dstatespace}
\bibfield{author}{\bibinfo{person}{Ciyu Ruan}, \bibinfo{person}{Ruishan Guo}, \bibinfo{person}{Zihang Gong}, \bibinfo{person}{Jingao Xu}, \bibinfo{person}{Wenhan Yang}, {and} \bibinfo{person}{Xinlei Chen}.} \bibinfo{year}{2025}\natexlab{}.
\newblock \bibinfo{title}{PRE-Mamba: A 4D State Space Model for Ultra-High-Frequent Event Camera Deraining}.
\newblock
\showeprint[arxiv]{2505.05307}~[cs.CV]
\urldef\tempurl%
\url{https://arxiv.org/abs/2505.05307}
\showURL{%
\tempurl}


\bibitem[Grompone~von Gioi et~al\mbox{.}(2010)]%
        {4731268}
\bibfield{author}{\bibinfo{person}{Rafael Grompone~von Gioi}, \bibinfo{person}{Jeremie Jakubowicz}, \bibinfo{person}{Jean-Michel Morel}, {and} \bibinfo{person}{Gregory Randall}.} \bibinfo{year}{2010}\natexlab{}.
\newblock \showarticletitle{LSD: A Fast Line Segment Detector with a False Detection Control}.
\newblock \bibinfo{journal}{\emph{IEEE Transactions on Pattern Analysis and Machine Intelligence}} \bibinfo{volume}{32}, \bibinfo{number}{4} (\bibinfo{year}{2010}), \bibinfo{pages}{722--732}.
\newblock
\href{https://doi.org/10.1109/TPAMI.2008.300}{doi:\nolinkurl{10.1109/TPAMI.2008.300}}


\end{thebibliography}

\end{sloppypar}
\end{document}